\documentclass[10pt, a4paper]{article}

\usepackage[final]{lrec2026} 
\usepackage{times}
\usepackage{latexsym}
\usepackage{tikz}
\usetikzlibrary{positioning,shadows,arrows.meta}
\usepackage[T1]{fontenc}
\usepackage[utf8]{inputenc}
\usepackage{microtype}
\usepackage{inconsolata}
\usepackage{graphicx}
\usepackage{amsmath}
\usepackage{float}
\usepackage{booktabs}
\usepackage{tabularx}
\usepackage{siunitx}
\usepackage[most]{tcolorbox}
\usepackage{xcolor}
\usepackage{amsmath,amssymb}
\usepackage[ruled,vlined]{algorithm2e}
\usepackage{placeins} 

\definecolor{arcblue}{HTML}{006699}
\definecolor{lightblue}{HTML}{E8F4F8}

\title{The Sufficiency-Conciseness Trade-off in LLM Self-Explanation from an Information Bottleneck Perspective}

\name{Ali Zahedzadeh, Behnam Bahrak}

\address{Tehran Institute for Advanced Studies, Khatam University, Tehran, Iran \\
	a.zahedzadeh@teias.institute, b.bahrak@teias.institute}

\abstract{
	Large Language Models increasingly rely on self-explanations, such as chain of thought reasoning, to improve performance on multi step question answering. While these explanations enhance accuracy, they are often verbose and costly to generate, raising the question of how much explanation is truly necessary. In this paper, we examine the trade-off between sufficiency, defined as the ability of an explanation to justify the correct answer, and conciseness, defined as the reduction in explanation length. Building on the information bottleneck principle, we conceptualize explanations as compressed representations that retain only the information essential for producing correct answers. To operationalize this view, we introduce an evaluation pipeline that constrains explanation length and assesses sufficiency using multiple language models on the ARC Challenge dataset. To broaden the scope, we conduct experiments in both English, using the original dataset, and Persian, as a resource-limited language through translation. Our experiments show that more concise explanations often remain sufficient, preserving accuracy while substantially reducing explanation length, whereas excessive compression leads to performance degradation.
	\\ \newline  }

\begin{document}
	
	\maketitleabstract
	
	\section{Introduction}
	
	When people answer complex exam questions, they are often expected not only to select the correct option but also to demonstrate their reasoning. Large Language Models (LLMs) are increasingly encouraged to do the same. Recent prompting strategies show that generating self-explanations most prominently chain of thought reasoning can substantially improve performance on reasoning intensive tasks \cite{wei2022chain}. Instead of directly predicting an answer, the model first produces intermediate reasoning steps, which increases the likelihood of success.
	
	Over time, several techniques have been proposed to refine this approach. Self-explanation prompting has been shown to enhance comprehension in dialogue understanding tasks \cite{gao2024selfexplanation}, while models are also capable of producing spontaneous self-explanations in general reasoning scenarios \cite{huang2023llmsexplain}. These advances underline the value of explanations, not only for accuracy but also for transparency and interpretability.
	
	However, more explanation is not always better. Models frequently generate verbose, repetitive, and sometimes misleading reasoning traces. Prior studies have warned that self-explanations are not guaranteed to faithfully reflect a model’s actual decision process \cite{madsen2024faithful,lyu2024faithfulsurvey}. Long chains also increase latency and computational cost, while many reasoning steps are unnecessary to reach the correct answer. This raises a fundamental question: How concise can an explanation be while still sufficient to justify the answer?
	
	From a theoretical perspective, this question connects to the Information Bottleneck (IB) framework, originally introduced by Tishby et al. \cite{tishby2000informationbottleneckmethod}. The IB principle formulates learning as finding compressed representations that discard irrelevant information while preserving what is predictive of the target. This framework was later extended to deep learning \cite{tishby2015deeplearningib,saxe2018ibtheory} and generalized in more recent analyses that relate compression to generalization and interpretability \cite{kawaguchi2023ibhelp,westphal2025generalizedib}. From this viewpoint, a good explanation should be a minimal sufficient representation retaining just enough information to justify the prediction.
	
	Recent work has begun to explore this balance between sufficiency and conciseness. Bassan et al. \cite{bassan2025explainbriefly} propose generating concise yet sufficient explanations through self-explaining neural architectures, while Bharti et al. \cite{bharti2024sufficientnecessary} and Amoukou and Brunel \cite{amoukou2022sufficientexplanations} formalize sufficiency and necessity as distinct dimensions of interpretability. Still, a systematic, large-scale investigation of the sufficiency–conciseness trade-off in LLM-generated self-explanations is limited, particularly across different languages and reasoning settings.
	
	In this work, we address this gap with an empirical study of explanation sufficiency under compression. We progressively constrain the length of explanations generated to answer questions in the ARC dataset \citelanguageresource{clark2018arc}, focusing on the ARC-Challenge subset, which requires multi-step reasoning beyond surface retrieval., in both English and Persian settings. Persian is included as a resource-limited language to broaden the evaluation scope. Sufficiency is assessed with a probe LLM model (Qwen 1.7B), while conciseness is measured by explanation-length reduction.
	
	Our contributions are as follows:
	\begin{itemize}
		\item We formalize sufficiency and conciseness as complementary dimensions of explanation quality, grounded in the Information Bottleneck perspective \cite{tishby2000informationbottleneckmethod}.
		\item We propose a general evaluation pipeline to test explanation sufficiency under progressive length constraints.
		\item We conduct the first bilingual study of explanation sufficiency, in English and Persian, identifying length thresholds for efficient yet sufficient self-explanations in large language models.
	\end{itemize}
	
	By quantifying the trade-off between sufficiency and conciseness, our work advances the study of explanation aware reasoning and provides practical insights for designing more efficient and trustworthy language models.

	\section{Related Work}
	
	\subsection{Explainable AI and Attribution Methods}
	Early work in explainable artificial intelligence focused on interpreting predictions of black-box models through feature attributions and visualization techniques. LIME \cite{ribeiro2016lime} and SHAP \cite{lundberg2017shap} introduced post-hoc local explanation frameworks that approximate complex models with interpretable surrogates or Shapley-based attributions. Integrated Gradients \cite{sundararajan2017ig} provided a theoretically grounded method for attributing deep network predictions, while saliency maps \cite{simonyan2014saliency} and attention mechanisms \cite{NIPS2017_3f5ee243,bahdanau2016attention,wiegreffe2019attention} visualized input contributions in neural models. These approaches typically rely on gradient access or model internals. In contrast, our method evaluates explanation sufficiency in a black-box setting requiring only model outputs which is crucial for evaluating proprietary large language models (LLMs) accessible solely via APIs.
	
	\subsection{Chain of Thought and Self-Explanations}
	Prompting strategies such as Chain-of-Thought (CoT) reasoning \cite{wei2022chain} and zero-shot reasoning \cite{kojima2022letsthink} demonstrate that eliciting intermediate reasoning steps improves performance on complex tasks. Subsequent research refined this idea through self-consistency decoding \cite{wang2023selfconsistencyimproveschainthought}, least-to-most prompting \cite{zhou2023leasttomostpromptingenablescomplex}, self-ask decomposition \cite{press-etal-2023-measuring}, and concise reasoning chains that perform comparably to verbose ones \cite{renze2024concise}, which enhance reasoning reliability. Other approaches, such as ReAct prompting \cite{yao2023react}, integrate reasoning with tool use, while reflective prompting \cite{liu2025selfreflection} encourages the model to critique and revise its own answers. Surveys such as \cite{zhang2024cot_survey} summarize these developments. Collectively, these works establish that explicit reasoning boosts accuracy and interpretability but they do not address how minimal an explanation can be while remaining sufficient.
	
	\subsection{Faithfulness and Sufficiency of Explanations}
	The faithfulness of model explanations has long been a central concern. ERASER \cite{deyoung2020eraser} formalized metrics such as sufficiency and comprehensiveness for evaluating rationalized NLP models, while Jacovi and Goldberg \cite{jacovi2020faithful} clarified the conceptual distinction between interpretability and faithfulness. Hase and Bansal \cite{hase-bansal-2022-models} proposed frameworks for analyzing when models truly learn from explanations. More recent studies have extended these ideas to LLM self-explanations: Madsen et al. \cite{madsen2024faithful} demonstrated that self-generated reasoning is often unfaithful to internal model behavior, and Huang et al. \cite{huang2023llmsexplain} found that explanations may be verbose, redundant, or misleading. These findings motivate direct evaluation of sufficiency whether explanations enable accurate prediction independent of surface plausibility.
	
	\subsection{Conciseness and Length Control}
	Renze and Guven \cite{renze2024concise} showed that concise reasoning chains can perform comparably to verbose ones, and Xu et al. \cite{xu2025chaindraftthinkingfaster} proposed “Chain of Draft,” generating short but informative reasoning traces. Gu et al. \cite{gu-etal-2025-length} developed a black-box iterative framework for precise length control without model modification, and Wu et al. \cite{wu2024moreisless} observed that excessively long explanations can even degrade reasoning quality. Despite this growing interest, no prior study has systematically quantified the sufficiency–conciseness trade-off in LLM self-explanations.
	
	\subsection{Information Bottleneck and Explanations}
	The Information Bottleneck (IB) framework \cite{tishby2000informationbottleneckmethod} views learning as compressing input representations while preserving task-relevant information. Later developments in deep learning \cite{tishby2015deeplearningib,saxe2018ibtheory,kawaguchi2023ibhelp,westphal2025generalizedib} extended this idea to representation learning. In the context of explainability, IB has inspired methods that treat explanations as compressed yet sufficient rationales \cite{paranjape2020ib,li-etal-2023-explanation}. Our study builds upon this perspective, viewing concise self-explanations as information-efficient justifications that retain predictive sufficiency.
	
	\subsection{Multilingual Self-Explanations}
	Recent research has explored reasoning and explanation transfer across languages. Shi et al. \cite{shi2022languagemodelsmultilingualchainofthought} showed that LLMs can generalize Chain-of-Thought prompting to multiple languages, while Barua et al. \cite{barua2025longcot} examined multilingual reasoning with longer chains. Surveys such as Ponti et al. \cite{ponti2023xling_survey} discuss persistent gaps in cross-lingual explanation fidelity. However, empirical work on sufficiency and conciseness in non-English contexts remains limited. Our bilingual experiments on the ARC dataset \citelanguageresource{clark2018arc} fill this gap by comparing English and Persian explanations under progressive compression.

	\section{Methodology}
	\label{sec:methodology}
	
	\subsection{Theoretical Background}
	Our study builds on the Information Bottleneck principle \cite{tishby2000informationbottleneckmethod}, which frames learning as the problem of compressing input representations $X$ into a bottleneck variable $Z$ while preserving information relevant to the target $Y$. The general objective of the information bottleneck is
	\begin{equation}
		\max_Z \; I(Z;Y) - \beta I(X;Z),
	\end{equation}
	where $I(\cdot;\cdot)$ denotes mutual information and $\beta$ is a balance parameter. In this framework, sufficiency corresponds to maximizing $I(Z;Y)$ so that the explanation $Z$ retains information required to justify the correct answer $Y$, while conciseness corresponds to minimizing $I(X;Z)$ so that the explanation does not redundantly encode irrelevant parts of the input $X$. Verbose explanations increase $I(X;Z)$ without necessarily improving $I(Z;Y)$, whereas overly short explanations risk losing sufficiency. Our approach operationalizes this trade off by constraining the length of explanations and analyzing whether sufficiency is preserved under these constraints.
	
	Directly computing the mutual information terms in the Information Bottleneck objective is intractable for large language models. This is primarily due to their black-box nature, which prevents access to the internal probability distributions required for such calculations. Therefore, we use practical proxy metrics to approximate the trade-off. We define \emph{Sufficiency} as the probability the model assigns to the correct answer given an explanation, and \emph{Conciseness} as the reduction in the explanation's length. This allows us to empirically evaluate the balance between generating informative and brief explanations.
	
	\subsection{Constrained Explanation Generation}
	Let $Z$ denote the full length explanation generated for a given question. To study conciseness, we prompt the model to regenerate its explanation under explicit word length constraints. For each constraint level $v \in \{10,20,\dots,90\}$, we instruct the model to produce an explanation $Z_v$ whose length is at most $(1 - v/100)$ fraction of the length of $Z$. For example, if $Z$ contains 50 words, then at the 20 percent constraint the model is required to produce an explanation of no more than 40 words. The unconstrained explanation is denoted $Z_0 := Z$. This procedure ensures that conciseness is enforced directly during generation rather than by post hoc trimming.
	
	\subsection{Sufficiency Evaluation}
	Evaluation is performed using an asymmetric setup: while multiple models generate explanations, a single fixed scorer model $M$ is employed to assess sufficiency across all conditions, ensuring comparability. For each constrained explanation $Z_v$, we construct a prompt $P_v$ comprising the question $Q$, the explanation $Z_v$, and the set of answer options $\mathcal{O} = \{A,B,C,D\}$. The scorer model outputs a probability distribution
	\begin{equation}
		p(o \mid P_v) \quad \text{for each } o \in \mathcal{O}.
	\end{equation}
	Let $y$ denote the gold-standard answer. We define sufficiency as the probability assigned by the scorer to the correct answer given the explanation:
	\begin{equation}
		\text{Sufficiency}(Z_v) = p(y \mid P_v).
	\end{equation}
	This metric directly quantifies the extent to which the explanation supports the correct answer. For reference, we also evaluate a baseline prompt $P_{noexp}$ excluding any explanation. This comparison reveals the increase in the scorer's confidence attributable to explanatory content, while sufficiency itself remains defined independently as $p(y \mid P_v)$.
	
	To prevent answer leakage, generated explanations are post-processed: any explicit mentions of answer option letters (A, B, C, D) or verbatim copies of option texts are replaced with a [MASK] symbol. This ensures that sufficiency measures genuine explanatory reasoning rather than superficial cues.
	
	\begin{algorithm}[ht!]
		\caption{Computing Sufficiency with a Fixed Scorer}
		\label{alg:sufficiency}
		\KwIn{Dataset $\mathcal{D}$ of items $(Q,\mathcal{O},y,Z_v)$ with $\mathcal{O}=\{A,B,C,D\}$; fixed scorer model $M$}
		\KwOut{Per-item sufficiency scores $\{\mathrm{Sufficiency}(Z_v)\}_v$ and their mean}
		
		\SetKwFunction{Score}{ScoreOptions}
		\SetKwProg{Fn}{Function}{:}{}
		
		\Fn{\Score{$M$, $P$}}{
			\ForEach{$o \in \mathcal{O}$}{
				$\text{lp}[o] \leftarrow$ sum of token log-probabilities that $M$ assigns to the string \texttt{"\space o"} conditioned on $P$
			}
			\Return $\mathrm{softmax}(\text{lp})$ \tcp*{distribution $p(o\mid P)$ over $\mathcal{O}$}
		}
		
		Initialize list $S \leftarrow [\,]$\;
		
		\ForEach{$(Q,\mathcal{O},y,Z_v) \in \mathcal{D}$}{
			$P_v \leftarrow$ prompt formed from $(Q, Z_v, \mathcal{O})$ with suffix \texttt{"The answer is "}\;
			$\mathbf{p} \leftarrow \Score(M, P_v)$ \tcp*{$\mathbf{p}[o]=p(o\mid P_v)$}
			$S.\mathrm{append}\!\left(\mathbf{p}[y]\right)$ \tcp*{$\mathrm{Sufficiency}(Z_v)=p(y\mid P_v)$}
		}
		
		\textbf{return} $S$, $\displaystyle \frac{1}{|S|}\sum_{s \in S} s$\;
	\end{algorithm}
	
	\noindent We also report the dataset-level mean sufficiency,
	\[
	\overline{\mathrm{Sufficiency}} \;=\; \frac{1}{|\mathcal{D}|}\sum_{(Q,\mathcal{O},y,Z_v)\in\mathcal{D}} p(y\mid P_v).
	\]

	\subsection{Conciseness Measurement}
	Conciseness is measured by the relative reduction in explanation length enforced during generation. At each constraint level $v$, the explanation $Z_v$ is shorter by $v$\% compared to the unconstrained explanation $Z_0$. By evaluating sufficiency across all levels, we can empirically analyze the trade-off between conciseness and sufficiency predicted by the information bottleneck framework.
	
	\subsection{Pipeline Summary}
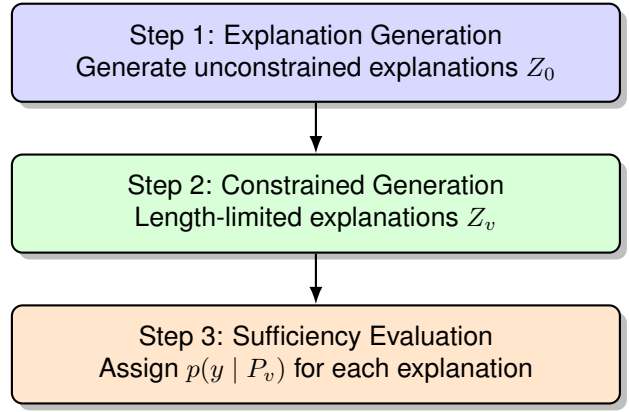
\begin{figure}[h]
		\centering
		\begin{tikzpicture}[node distance=2cm, >=Latex, thick]
			
			\tikzstyle{process} = [rectangle, draw, rounded corners,
			minimum width=8cm, minimum height=1.3cm, align=center,
			fill=blue!15, drop shadow]
			
			\tikzstyle{data} = [rectangle, draw, rounded corners,
			minimum width=8cm, minimum height=1.3cm, align=center,
			fill=green!15, drop shadow]
			
			\tikzstyle{result} = [rectangle, draw, rounded corners,
			minimum width=8cm, minimum height=1.3cm, align=center,
			fill=orange!20, drop shadow]
			
			\node[process] (gen) {Step 1: Explanation Generation \\ 
				Generate unconstrained explanations $Z_0$};
			\node[data, below of=gen] (con) {Step 2: Constrained Generation \\ 
				Length-limited explanations $Z_v$};
			\node[result, below of=con] (eval) {Step 3: Sufficiency Evaluation \\ 
				Assign $p(y \mid P_v)$ for each explanation};
			
			\draw[->] (gen) -- (con);
			\draw[->] (con) -- (eval);
			
		\end{tikzpicture}
		\caption{Pipeline of the proposed methodology. Explanations are generated by LLMs, then regenerated under explicit length constraints, and finally evaluated for sufficiency using a fixed scorer model.}
		\label{fig:pipeline}
	\end{figure}

	The complete pipeline consists of three stages as illustrated in Figure~\ref{fig:pipeline}. First, explanations are generated by multiple models under a shared prompting scheme. Second, constrained versions of each explanation are generated at predefined reduction levels through explicit length control in the prompt. Third, the fixed scorer model evaluates all explanations and returns sufficiency values defined as probabilities assigned to the correct answer. Baseline runs without explanations are included for comparison. Throughout the pipeline, structured logging and storage in the Comma-Separated Values (CSV) format ensure reproducibility and enable systematic quantitative analysis.
	
	\section{Experimental Setup}
	
	\subsection{Dataset}
	We conducted all experiments on the ARC Challenge dataset \citelanguageresource{clark2018arc}, a benchmark of multiple choice science questions designed to test advanced reasoning capabilities. The dataset contains 7,787 questions in total, divided into an Easy Set and a Challenge Set. The Challenge Set, which we use in this study, consists of 2590 four choice questions that are known to be difficult for retrieval based and co-occurrence based algorithms. Each question requires non-trivial reasoning rather than surface level matching. 
	
	To broaden the scope of evaluation, we also create a Persian version of the dataset by translating the original English questions and answer options. Persian is included as a resource-limited language to evaluate the robustness of explanation sufficiency across linguistic settings.

	\subsection{Models and Prompting}
	We evaluated seven large language models from diverse providers as explanation generators:
	GPT-4o-mini~\cite{hurst2024gpt4o} (OpenAI), Claude~3~Haiku~\cite{anthropic2024claude35} (Anthropic), Llama~4~Scout~\cite{meta2025llama4} (Meta, 109 billion total parameters with 17 billion active in MoE architecture~\cite{sanseviero2023moe,mu2025comprehensivesurveymixtureofexpertsalgorithms}), Gemini~2.0~Flash~\cite{geminiteam2024gemini15} (Google), Cohere~Command~R~\cite{cohere2024commandr} (Cohere, 35 billion parameters), DeepSeek-V3.1~\cite{deepseekai2025deepseekv3} (DeepSeek, 671 billion total parameters with 37 billion active in MoE architecture~\cite{sanseviero2023moe,mu2025comprehensivesurveymixtureofexpertsalgorithms}), and Mistral~Small~3.1~\cite{jiang2024mixtral} (Mistral, 24 billion parameters).
	These models were selected for their cost-effectiveness, as the evaluation required generating outputs in multiple length variants, which would have incurred substantial expenses if more advanced models had been employed. The variation in parameter counts, including dense and Mixture-of-Experts (MoE) architectures, further supports efficient inference while maintaining diverse capabilities.
	All models were accessed through the unified OpenRouter~API~\cite{openrouter2025},
	which standardizes query parameters and temperature control across providers.
	A single fixed scorer model, Qwen3~1.7B~\cite{qwen3},
	was used to evaluate explanation sufficiency, ensuring consistent judgments across all generated explanations.
	All models are prompted with a uniform template. Each model is instructed to produce a final prediction in the form of one option among A, B, C, or D, and to accompany it with a natural language explanation. For length constrained variants, models are instructed to regenerate their explanations under explicit word length limits defined as percentages of the original explanation length. Prompt templates are provided in the final version of this paper.
	
	\subsection{Implementation Details}
	We evaluate on the full 2590 question ARC Challenge dataset. For each generated explanation, we create multiple constrained versions corresponding to nine reduction levels ranging from 10 percent to 90 percent of the original explanation's length. To prevent answer leakage, we apply masking to all explanations prior to evaluation. If an explanation explicitly mentions option labels (A, B, C, D) or copies answer text verbatim, such tokens are replaced with a [MASK] symbol. The scorer model then evaluates sufficiency for each explanation by assigning probabilities to the four answer options. All runs are executed in Python with structured logging and storage in CSV format to facilitate reproducibility.

	\section{Results}
	
	In this section, we present our experimental findings on the ARC-Challenge dataset in both English and Persian. We evaluate the quality of model-generated explanations under progressively tighter length constraints using four complementary metrics that jointly capture task performance, explanatory sufficiency, and semantic preservation:
	
	\begin{enumerate}
		\item \textbf{Accuracy:} the proportion of correctly predicted answers produced by the model. This metric reflects the overall task performance and serves as a direct measure of reasoning success.
		
		\item \textbf{Sufficiency:} the probability assigned by the fixed scorer model to the correct answer, given the explanation. This metric quantifies how effectively an explanation enables the correct prediction, independent of surface plausibility.
		
		\item \textbf{Embedding Similarity:} the cosine similarity between sentence-level embeddings of the base explanation and its compressed variant. This measures semantic alignment and helps determine whether shorter explanations retain the same meaning.

	\end{enumerate}
	

	Together, these three perspectives allow us to systematically investigate the trade-off between conciseness and explanatory adequacy.
	
	\subsection{Baseline Performance of the Scorer Model}
	\label{subsec:baseline-scorer}
	
	Before analyzing the effect of explanations on sufficiency, we first report the
	\textit{baseline performance} of the scorer model itself. In this setting, the model only receives the question and the answer options, without any explanation. We measure two metrics:
	
	\begin{itemize}
		\item \textbf{Baseline Accuracy:} the percentage of cases where the selected option by the scorer model matches the gold answer.
		\item \textbf{Baseline Sufficiency:} the average probability assigned to the gold option by the scorer model in the absence of explanations.
	\end{itemize}
	
	This baseline provides a reference point for interpreting subsequent improvements or declines when explanations are added or shortened. The results are presented in Table~\ref{tab:baseline-scorer}.

	\begin{table}[H]
		\centering
		\caption{Baseline accuracy (\%) and sufficiency (\%) of the scorer model without explanations.}
		\label{tab:baseline-scorer}
		\begin{tabular}{lcc}
			\toprule
			\textbf{Language} & \textbf{Accuracy} & \textbf{Sufficiency} \\
			\midrule
			English & 71.17 & 80.71 \\
			Persian & 47.85 & 72.82 \\
			\bottomrule
		\end{tabular}
	\end{table}

	\subsection{Embedding Similarity Analysis}
	\label{subsec:similarity}
	\begin{figure*}[t]
		\centering
		\includegraphics[width=0.48\linewidth]{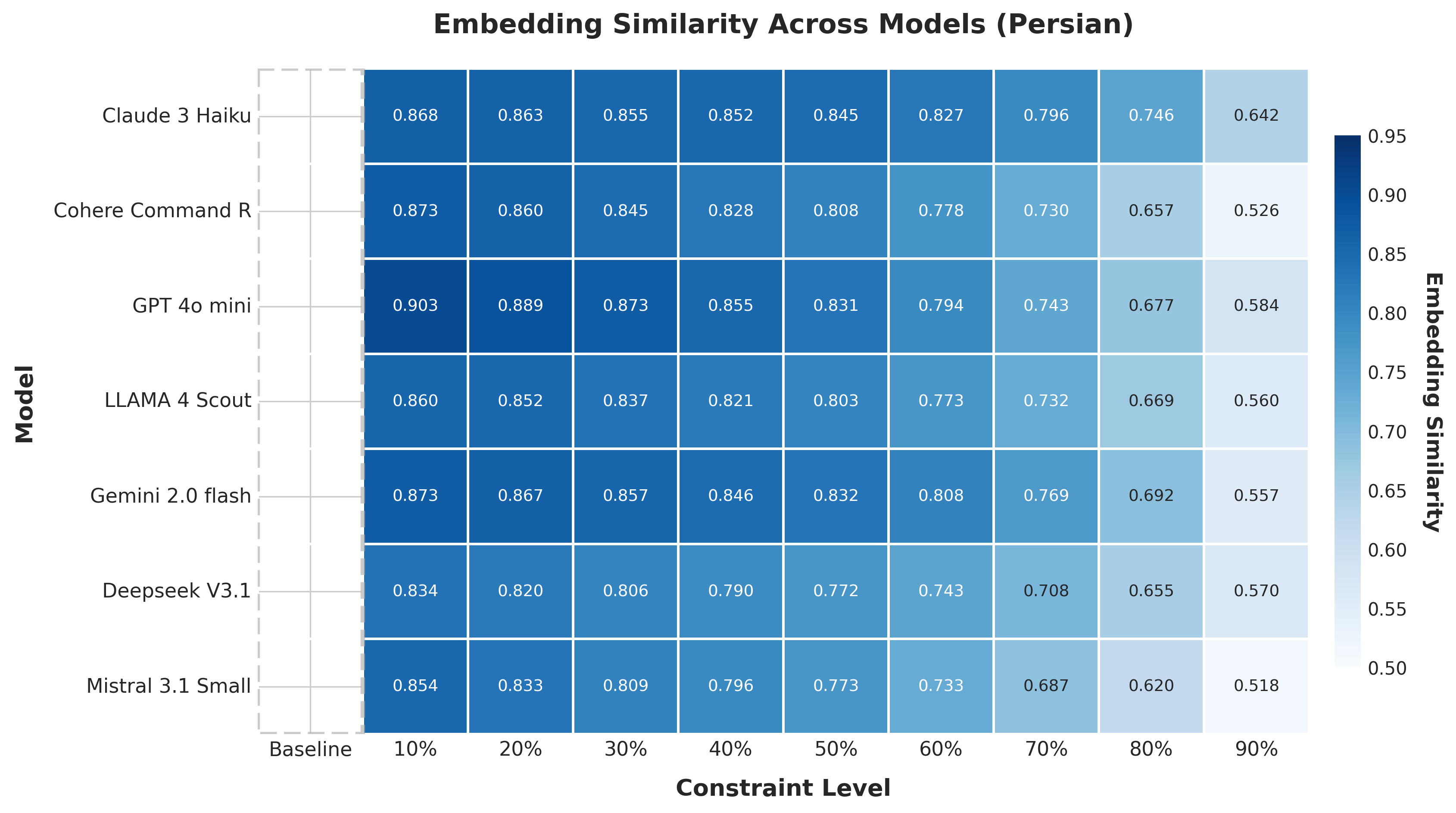}
		\includegraphics[width=0.48\linewidth]{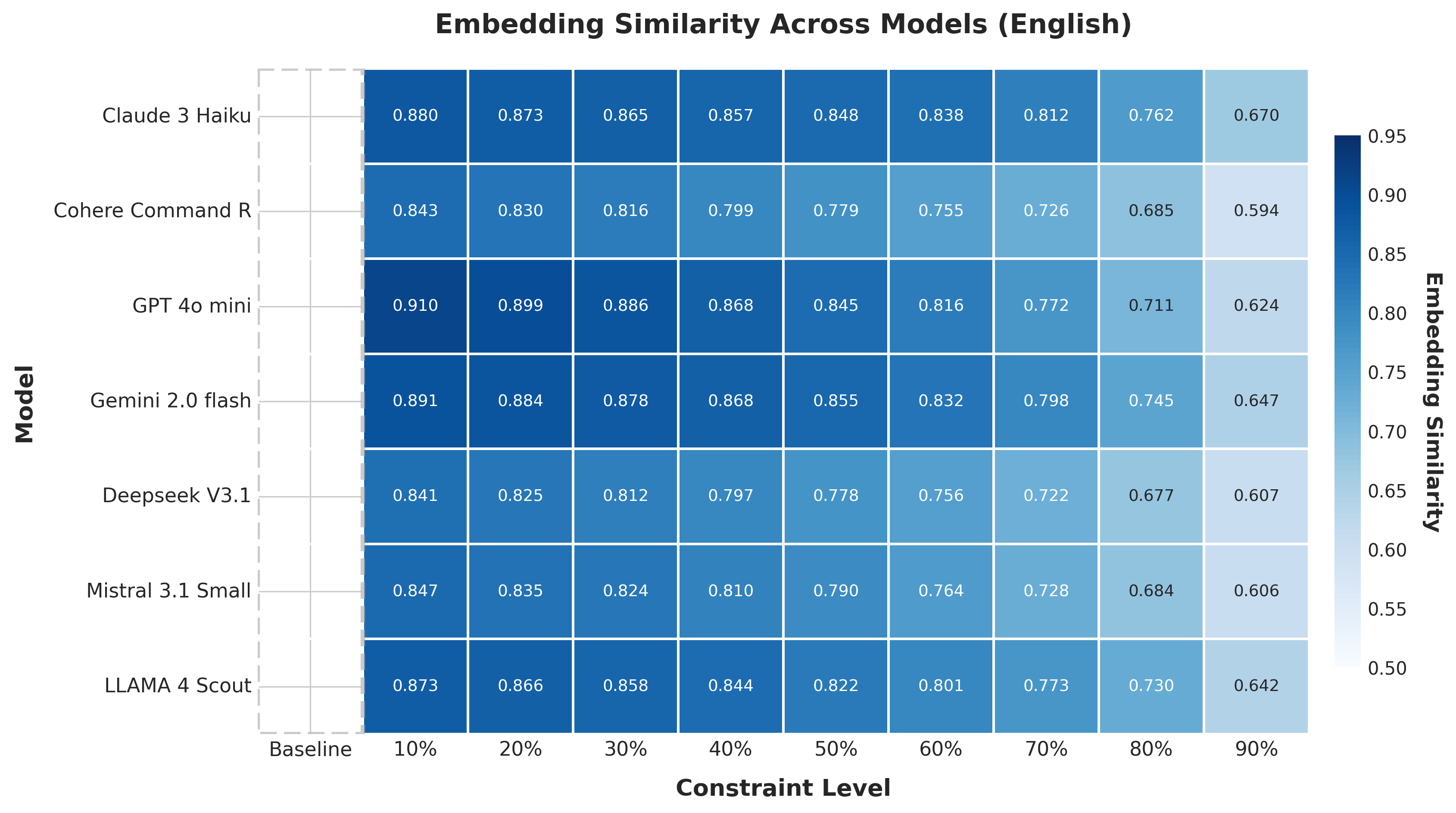}
		\caption{Average embedding similarity between base and length-constrained explanations across models and constraint levels, shown as heatmaps for Persian (left) and English (right). Darker shades indicate stronger semantic preservation.}
		\label{fig:similarity-heatmap}
	\end{figure*}
	Beyond accuracy and sufficiency, we further analyze the \textit{semantic stability} of explanations under progressive compression. This analysis measures how closely shortened explanations preserve the original meaning, thereby quantifying the extent of semantic drift caused by conciseness constraints.
	For each question, we encode both the base explanation $Z_0$ and its length-limited variants ($Z_{10}$–$Z_{90}$) using the \texttt{Qwen/Qwen3-Embedding-0.6B} model. We then compute the cosine similarity between the embedding vectors of each pair and average the results over the dataset. The resulting score lies in the range $[0,1]$, where higher values indicate stronger preservation of meaning.
	To visualize cross-model and cross-language patterns, we report the mean similarity across all constraint levels as a heatmap (Figure~\ref{fig:similarity-heatmap}). Each cell represents the average embedding similarity between the unconstrained explanation and its compressed counterpart for a specific model and constraint level. Darker shades correspond to higher semantic overlap.
	Overall, the heatmaps reveal a consistent downward gradient from left to right, indicating semantic drift under compression. However, the rate of degradation varies across models and languages. In English, models such as Claude 3 Haiku and Gemini 2.0 Flash maintain relatively high similarity scores (above 0.80) at moderate compression levels, suggesting that their explanations are more information-dense. In contrast, Persian explanations degrade more sharply after the 40\% level, reflecting both linguistic complexity and lower model exposure to Persian training data. Larger models like DeepSeek-V3.1 (671 billion total parameters) exhibit sustained similarity above 0.80 in English, potentially due to greater information redundancy.
	Interestingly, semantic similarity does not always align with sufficiency: even when explanations remain semantically close to the original ($\mathrm{sim}\ge0.85$), their sufficiency scores can drop noticeably. This observation indicates that beyond lexical or semantic preservation, structural and causal cues in the explanation are also essential for maintaining reasoning effectiveness.
	
	\subsection{Overall Accuracy and Sufficiency}
	Table~\ref{tab:overall-z0} reports accuracy and sufficiency scores for all seven LLMs using full explanations ($Z_0$) on the ARC Challenge dataset. Results are shown for both the original English setting and the Persian translated version. Across all models, explanations substantially improve performance compared to the no-explanation baseline (presented in Table~\ref{tab:baseline-scorer}), confirming that self-explanations provide critical support for reasoning. However, we also observe a consistent gap between English and Persian performance, reflecting the increased difficulty of reasoning in a resource-limited language.
	\begin{table*}[t]
		\centering
		\caption{Accuracy and sufficiency (\%) with full explanations ($Z_0$) for all models on ARC Challenge in English and Persian.}
		\label{tab:overall-z0}
		\begin{tabular}{lcccc}
			\toprule
			\textbf{Model} & \textbf{English Acc} & \textbf{English Suff} & \textbf{Persian Acc} & \textbf{Persian Suff} \\
			\midrule
			GPT-4o-mini & 89.34 & 86.18 & \textbf{82.60} & \textbf{79.18} \\
			Claude 3 Haiku & 86.20 & 83.05 & 78.38 & 74.77 \\
			LLaMA 4 Scout & 89.76 & \underline{86.76} & 79.62 & 75.81 \\
			Gemini 2.0 Flash & 89.77 & 85.74 & \underline{82.06} & \underline{78.08} \\
			Cohere Command R & 85.00 & 81.81 & 69.23 & 66.12 \\
			DeepSeek V3.1 & \textbf{90.74} & \textbf{87.32} & 80.24 & 76.60 \\
			Mistral Small 3.1 & \underline{90.00} & 86.18 & 75.04 & 71.63 \\
			\bottomrule
		\end{tabular}
	\end{table*}
	As seen in Table~\ref{tab:overall-z0}, all models achieve higher accuracy and sufficiency with explanations, with DeepSeek V3.1 and Mistral Small 3.1 performing strongest in English, while GPT-4o-mini and Gemini 2.0 Flash lead in Persian. Persian results consistently lag behind English, highlighting the challenges of applying explanation-based reasoning in low-resource linguistic settings. Notably, larger models such as DeepSeek-V3.1 (671 billion total parameters) excel in English, whereas smaller ones like GPT-4o-mini (8 billion parameters) demonstrate resilience in Persian, suggesting that parameter scale interacts with language-specific training.
	
	\subsection{Effect of Conciseness on Accuracy and Sufficiency}
	\label{subsec:conciseness}
\begin{figure}[t]
		\centering
		\includegraphics[width=0.98\linewidth]{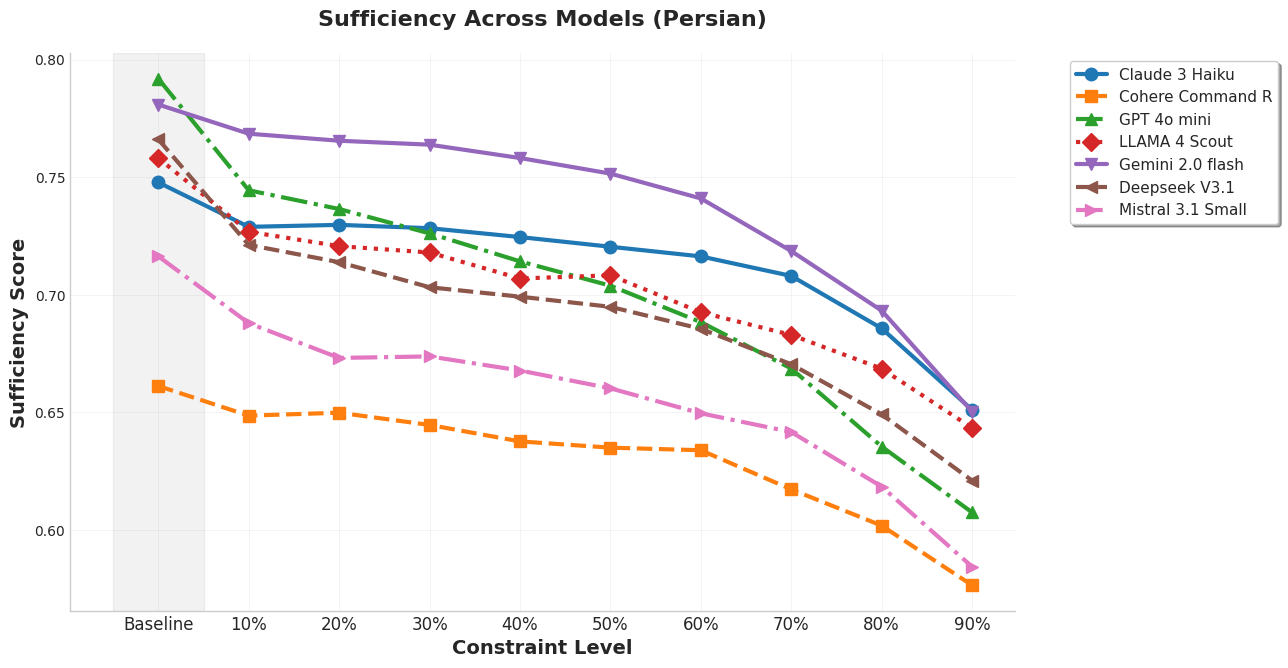}
		\includegraphics[width=0.98\linewidth]{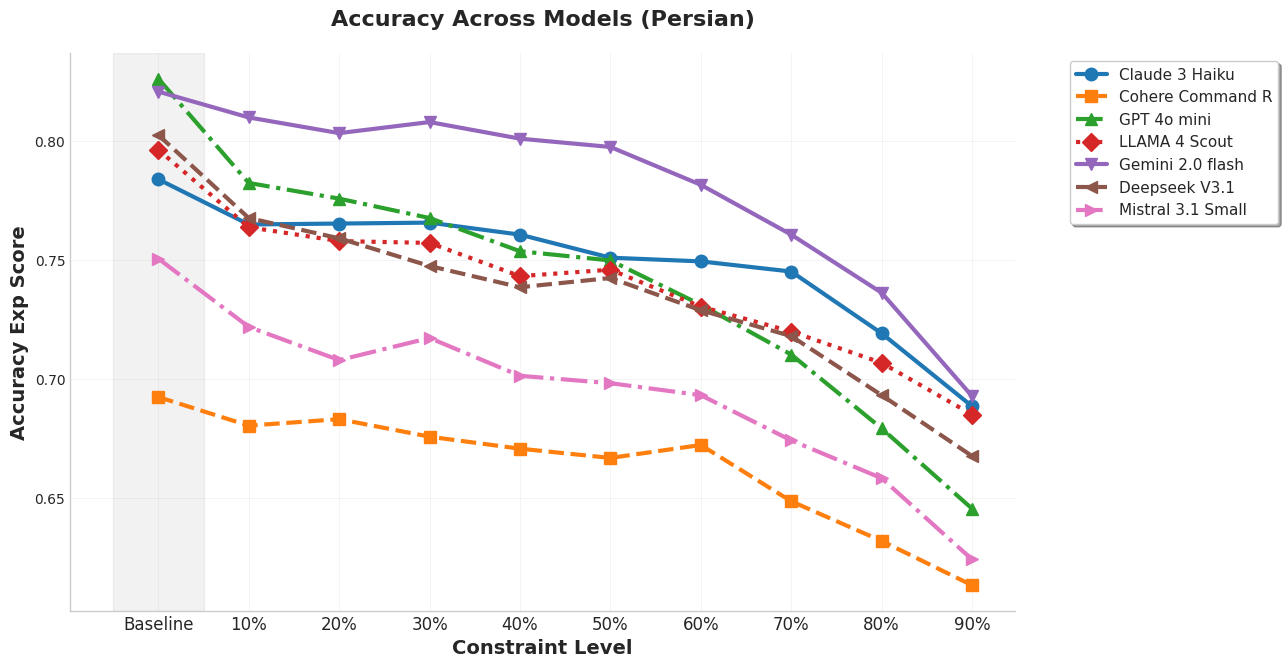}
		\caption{Sufficiency (up) and Accuracy (down) across explanation length constraints in Persian.}
		\label{fig:sufficiency-fa}
	\end{figure}

\begin{figure}[t]
		\centering
		\includegraphics[width=0.98\linewidth]{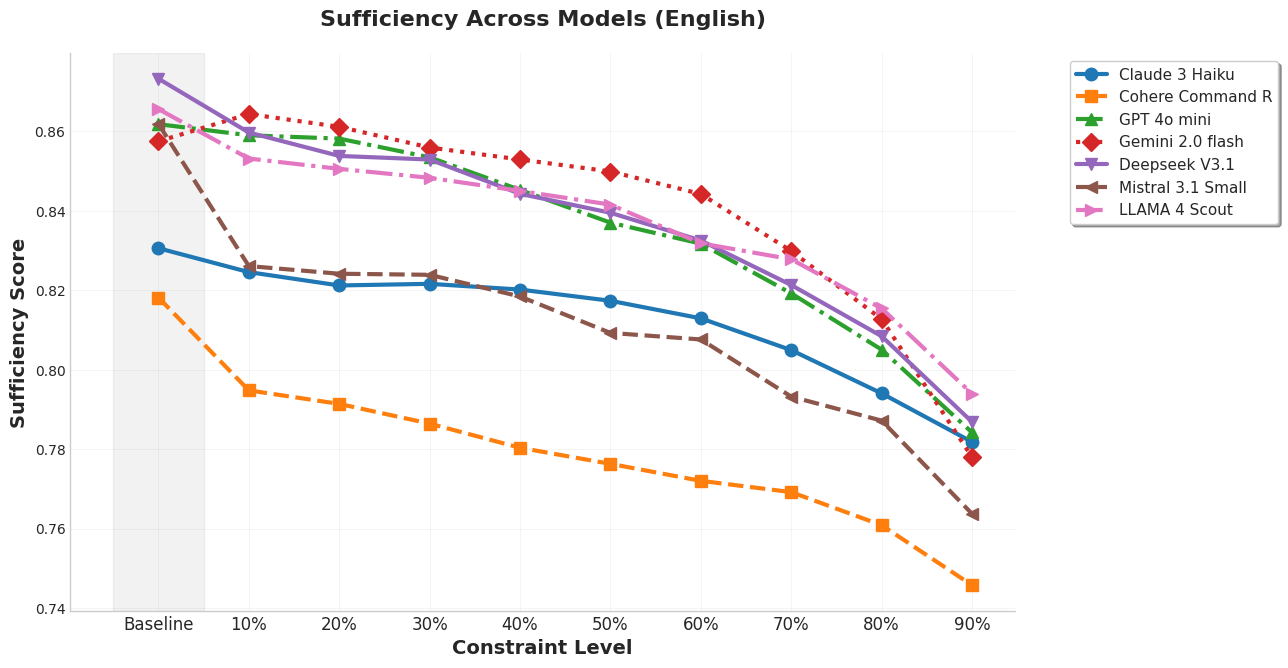}
		\includegraphics[width=0.98\linewidth]{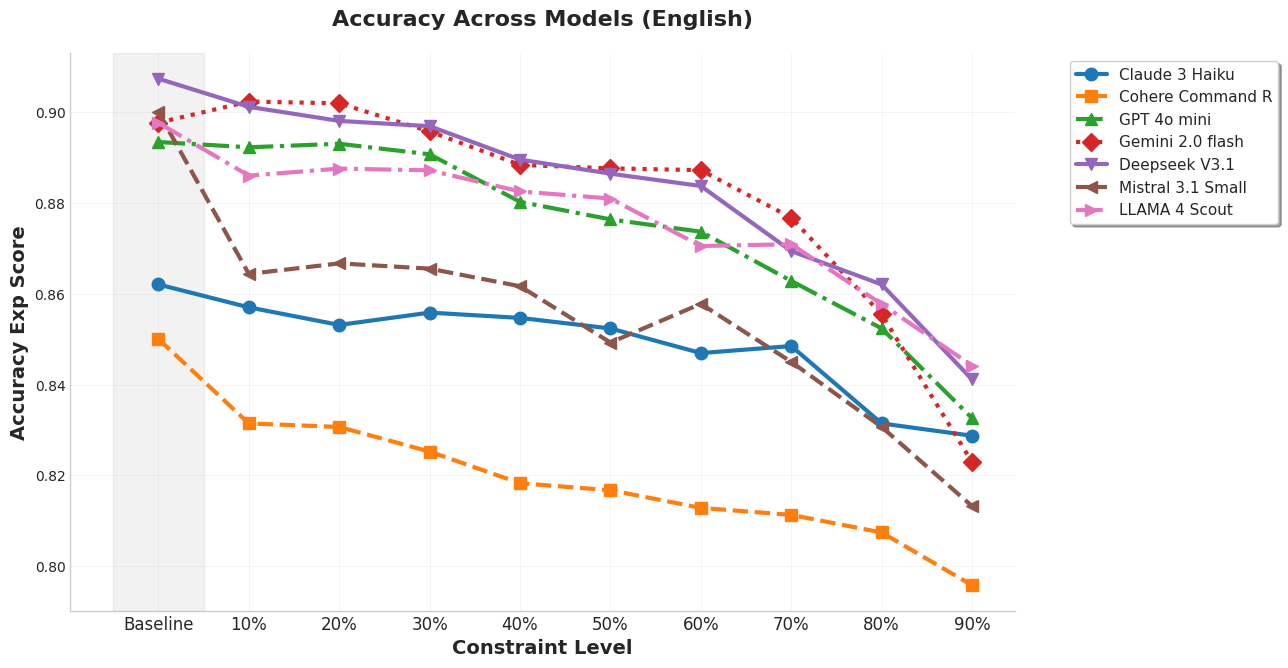}
		\caption{Sufficiency (up) and Accuracy (down) across explanation length constraints in English.}
		\label{fig:sufficiency-en}
	\end{figure}

	We next analyze how progressively shortening explanations impacts both task accuracy and sufficiency.
	Figures~\ref{fig:sufficiency-fa} and \ref{fig:sufficiency-en} present results across all generator models on the ARC Challenge dataset in Persian and English, respectively.
	Several consistent patterns emerge. First, both accuracy and sufficiency decrease as explanations are shortened, confirming that longer explanations generally provide more reliable reasoning support.
	Second, the degradation is steeper in Persian compared to English, reflecting the additional difficulty of generating effective concise explanations in a low-resource language.
	Third, model differences are evident: in English, DeepSeek V3.1 and GPT-4o-mini demonstrate greater robustness, maintaining accuracy scores above 0.84 even at 90\% constraint levels, whereas LLaMA 4 Scout and Mistral Small 3.1 exhibit sharper declines, dropping below 0.75. In Persian, Gemini 2.0 Flash and Claude 3 Haiku preserve performance more effectively under high compression, with sufficiency scores remaining above 0.66 at 90\% constraints, compared to steeper drops in models like Cohere Command R and Mistral Small 3.1. These variations suggest that differences in instruction-following capabilities, multilingual training, and alignment may contribute to enhanced robustness under conciseness constraints.
	\section{Conclusion}
	Our work examined the trade-off between explanation sufficiency and conciseness in large language models through the lens of the Information Bottleneck principle. 
	We proposed a systematic evaluation pipeline to constrain and assess self-explanations, applying it to both English and Persian versions of the ARC-Challenge dataset.
	Our findings reveal that explanations can often be substantially shortened while preserving sufficiency, indicating that many reasoning chains contain redundant steps. 
	This suggests that concise explanations not only reduce computational cost but also align more closely with human-like efficiency in reasoning.
	Future work will extend the framework to open-ended tasks, investigate automatic conciseness control, and explore sufficiency preservation under multilingual and multimodal conditions.

	\section{Ethical Considerations}
	All datasets and models employed in this research are publicly available and adhere to established ethical guidelines for AI research. The ARC-Challenge dataset comprises 2,590 genuine grade-school-level, multiple-choice science questions designed to advance question-answering methodologies, with no inclusion of personal, sensitive, or identifiable information. The Persian translation of the dataset was generated automatically and manually reviewed for fidelity, ensuring it contains no human-identifiable content or biases that could arise from mistranslation. All large language models were accessed via official APIs in full compliance with their respective usage policies, mitigating risks associated with unauthorized data handling. No human subjects participated in this study, thereby eliminating concerns related to informed consent or privacy. Additionally, we acknowledge potential environmental impacts from LLM inference and advocate for energy-efficient practices in future extensions of this work.

	\section{References}\label{sec:reference}
	\bibliographystyle{lrec2026-natbib}
	\bibliography{references}
	
	\section{Language Resource References}\label{lr:ref}
	\bibliographystylelanguageresource{lrec2026-natbib}
	\bibliographylanguageresource{languageresource}

\end{document}